\documentclass[11pt]{article}

\usepackage[preprint]{acl}

\usepackage{times}
\usepackage{latexsym}
\usepackage{xcolor}
\usepackage{soul}
\usepackage{todonotes}
\usepackage{cuted}     
\usepackage{tcolorbox}
\tcbuselibrary{skins}

\usepackage[T1]{fontenc}

\usepackage[utf8]{inputenc}

\usepackage{microtype}

\usepackage{inconsolata}

\usepackage{graphicx}
\usepackage{amssymb}
\usepackage{amsmath}
\usepackage{amsthm}
\newtheorem{proposition}{Proposition}

\newcommand{\grpol}{GRPO style }

\usepackage{booktabs}   
\usepackage{multirow}   
\usepackage{array}      

\newtcolorbox{myprop}[1][]{
    enhanced,                          
    colback=gray!5!white,              
    colframe=gray!70!black,            
    boxrule=0.5mm,                     
    arc=2mm,
    drop shadow={gray!40!white},       
    left=5pt, right=5pt,             
    top=5pt, bottom=5pt,             
    #1                                 
}
\title{On the Hidden Objective Biases of Group-based Reinforcement Learning}

\author{
  Aleksandar Fontana\textsuperscript{\rm 1,2}\thanks{Equal contribution.},
  Marco Simoni\textsuperscript{\rm 2,3}\footnotemark[1],
  Giulio Rossolini\textsuperscript{\rm 1},
  Andrea Saracino\textsuperscript{\rm 1}, 
  Paolo Mori\textsuperscript{\rm 2} \\
  \textsuperscript{1} Department of Excellence in Robotics and AI, TeCIP, Scuola Superiore Sant’Anna, Pisa \\
  \textsuperscript{2} Institute of Informatics and Telematics, National Research Council of Italy, Pisa \\
  \textsuperscript{3} National Doctorate on Artificial Intelligence, Sapienza Università di Roma\\
}

\begin{document}
\maketitle
\begin{abstract}

Group-based reinforcement learning methods, like Group Relative Policy Optimization (GRPO), are widely used nowadays to post-train large language models. Despite their empirical success, they exhibit structural mismatches between reward optimization and the underlying training objective.
In this paper, we present a theoretical analysis of \grpol methods by studying them within a unified surrogate formulation. This perspective reveals recurring properties that affect all the methods under analysis: (i) non-uniform group weighting induces systematic gradient biases on shared prefix tokens; (ii) interactions with the AdamW optimizer make training dynamics largely insensitive to reward scaling; and (iii) optimizer momentum can push policy updates beyond the intended clipping region under repeated optimization steps.
We believe that these findings highlight fundamental limitations of current approaches and provide principled guidance for the design of future formulations.

\end{abstract}

\section{Introduction}

Recent advances in Large 
Language Model (LLM) post-training have shown that reinforcement learning methods based on group-level feedback can effectively improve reasoning performance while avoiding the cost of explicit value-function estimation, as used in previous works \cite{rlhf,Tree-of-thoughts}.
Among these approaches, Group Relative Policy Optimization (GRPO) and related methods have gained widespread adoption due to their simplicity and scalability, and are now commonly used in post-training pipelines for reasoning-oriented models
\cite{shao2024deepseekmath, liu2025deepseek, zheng2025group, yu2025dapo}.

Despite their empirical success, \grpol methods rely on a surrogate
objective whose optimization dynamics remain only partially understood.
Several recent works have reported unexpected behaviors during training, including
length-related biases \cite{liu2025understanding}, sensitivity to formatting tokens
\cite{simoni2025gtpo}, reward hacking in multi-objective settings
\cite{ichihara2025mo}, and instability across different optimization regimes
However, these findings represent fragmented empirical observations, and a unified formal framework that systematically connects and further extends them to the surrogate objective’s implicit inductive biases is lacking.

This work offers a unified critical analysis of group-based optimization methods. We propose a general formulation of \grpol methods, showing ten recent approaches as special cases.
This view reveals shared issues, showing that the surrogate objective is often dominated by weighting schemes, regularization, and importance sampling, rather than by pure reward maximization. Building on this formulation, we identify three recurring properties of \grpol training dynamics: (i) we analyze token-level gradients to demonstrate that non-uniform weighting induces systematic biases on shared prefix tokens; (ii) 
we study the interaction with AdamW~\cite{loshchilov2017decoupled}, demonstrating that the training process remains invariant to global reward scaling across various scenarios; (iii) we show that optimizer momentum can drive policy updates beyond the intended clipping boundaries during multi-step optimization. 
Beyond empirical performance, our analysis offers theoretical insights exposing a divergence between the surrogate objective and the true training goal. By characterizing these dynamics, our findings provide the community with a reference for the design and interpretation of LLM post-training strategies.

\begin{figure*}[t]
\centering
\small

\begin{tcolorbox}[
    colback=gray!5,
    colframe=black!70,
    boxrule=0.6pt,
    arc=2pt,
    left=8pt, right=8pt,
    top=4pt, bottom=6pt,
    width=\textwidth,
    title=\centering \textbf{\grpol  Objective},
    fonttitle=\normalsize\bfseries
]
\begin{equation}
\mathcal{J}_{\text{GRPO-L}}(\theta) =
\mathbb{E}_{q,\{o_i\}} \Bigg[
\sum_{i=1}^G \left(
\sum_{t=1}^{|o_i|}\alpha_{i,t}
\min\Big(
s_{i,t}(\theta)\,A_i,\;
\operatorname{clip}\big(s_{i,t}(\theta), 1-\varepsilon_{low}, 1+\varepsilon_{up}\big)\,A_i
\Big)
\right)
- \beta R(\theta)
\Bigg]
\label{eq:grpol}
\end{equation}
\end{tcolorbox}
\end{figure*}
\section{Related Work}

Recent work has started to study the problems arising during \grpol post-training.
Several studies report optimization issues, like systematic biases
toward output length~\cite{liu2025understanding}.
Other analyses propose simple stabilization techniques, including masking strategies,
to improve robustness across different training regimes~\cite{mroueh2025revisiting}.
In multi-objective settings, GRPO has is vulnerable to reward
hacking, motivating the use of normalization-based mitigations~\cite{ichihara2025mo}.
Additional work focuses on issues that emerge at the token level: formatting tokens often dominate optimization~\cite{simoni2025gtpo}, and simple cues like sequence length can drive learning~\cite{xin2025surrogate}.
Clipping mechanisms used in PPO and GRPO have also been shown to introduce systematic
entropy biases~\cite{park2025clip}. Complementary to analyses of clipping and instability, SFPO introduces a reposition-before-update scheme to control off-policy drift induced by repeated inner updates~\cite{wang2025slow}.
Based on these observations, our work provides a unified analysis of why the
surrogate loss can be misleading, how shared prefixes bias token-level gradients, and
how optimizer dynamics interact with clipping under repeated updates. 



\normalsize
\section{Unified Formulation}

In the following, we introduce a generalized surrogate objective that serves as a unified framework for a broad class of recent group-based policy optimization methods, including GRPO (R1~\cite{shao2024deepseekmath} and v3.2~\cite{liu2025deepseek}), GSPO\footnote{We report GSPO-token, as it yields the same gradients and optimization trajectory as standard GSPO.}~\cite{zheng2025group}, GTPO~\cite{simoni2025gtpo}, DAPO~\cite{yu2025dapo}, CPPO~\cite{lin2025cppo}, Dr.~GRPO~\cite{liu2025understanding}, GPG~\cite{GPG}, CISPO~\cite{CISPO}, and GCPO~\cite{GCPO}. For a group of $G$ outputs $\{o_i\}_{i=1}^G$ generated from the same prompt $q$, the advantage $A_i$ for the $i$-th output is calculated by standardizing the reward $r_i$ against the group's distribution:

\begin{equation}
    A_i = r_i - \left(\frac{1}{G}\sum_{j=1}^G r_j\right)
    \label{eq:advantage}
\end{equation}

\normalsize
%
This advantage term drives the \textit{\grpol objective} (Eq.~\ref{eq:grpol}). $A_i$ usually determines the direction of the token-level policy updates weighting coefficients $\alpha_{i,t}$. Optimization typically involves $\mu$ gradient updates on a fixed group of samples, which progressively induces off-policy drift. To mitigate this, it is employed a token-level importance ratio 
\begin{equation}
    s_{i,t}(\theta) \propto
\frac{\pi_\theta(y_{i,t} \mid x, y_{i,<t})}
     {\pi_{\theta_{\text{old}}}(y_{i,t} \mid x, y_{i,<t})}
\end{equation}
clipped to $[1 - \varepsilon_{\text{low}},\, 1 + \varepsilon_{\text{up}}]$ following PPO~\cite{schulman2017proximal}.
Finally, a regularization term $R(\theta)$, generally the KL divergence from a reference policy weighted by $\beta$, is added for training stability. As detailed in Table~\ref{tab:algorithm_compact_double_short}, each method represents a distinct configuration of Eq.~\ref{eq:grpol} regarding the three core components: the weighting coefficients $\alpha_{i,t}$, the importance ratio $s_{i,t} (\theta)$, and the regularization term $R(\theta)$. Eq.~\ref{eq:grpol} acts strictly as an optimization mechanism, not as a performance metric. Since advantages are group-centered ($\sum_i A_i = 0$), the loss value does not exclusively reflect reward improvement. Instead, the loss magnitude is dominated by nuisance factors, like importance sampling fluctuations ($s_{i,t} \neq 1$) during multi-step updates. Consequently, the surrogate loss offers no monotonic or reliable signal for policy improvement and should not be used to monitor training progress \cite{achiam2018spinning} (see Appendix~\ref{app:loss_proxy} for formal analysis).

\begin{table*}[t]
    \centering
    \renewcommand{\arraystretch}{1.15} 
    \setlength{\tabcolsep}{3pt}      
    
    \caption{Instantiation of the unified objective in Eq.~\ref{eq:grpol} for representative \grpol methods. 
Weights $\alpha$, importance ratios $s_{i,t}(\theta)$, and regularization terms $R(\theta)$ are reported for each algorithm. The definitions are: $\alpha_i^S := \frac{1}{G \cdot |o_i| \cdot \sigma(r)}$, $I := \frac{\pi_\theta}{\pi_{\theta_{old}}}$, and $\mathcal{D}_{KL} := \frac{\pi_{ref}}{\pi_{\theta}} - \log \frac{\pi_{ref}}{\pi_{\theta}} - 1$. Unless otherwise specified, dependence on $(y_{i,t}\mid x, y_{i,<t})$ is implicit.}
    \label{tab:algorithm_compact_double_short}
    
    \resizebox{\textwidth}{!}{
    \begin{tabular}{ l c c c | l c c c }
        \toprule
        \textbf{Algorithm} & \textbf{$\alpha_{i,t}$} & \textbf{$s_{i,t}(\theta)$} & \textbf{$R(\theta)$} & 
        \textbf{Algorithm} & \textbf{$\alpha_{i,t}$} & \textbf{$s_{i,t}(\theta)$} & \textbf{$R(\theta)$} \\
        
        \cmidrule(r){1-4} \cmidrule(l){5-8} 
        
        GRPO R1 & $\alpha_i^S$ & $I$ & $\mathcal{D}_{KL}$ & 
        CPPO & $\alpha_i^S 1_{\{|A_i|>\gamma\}}$ & $I$ & $\mathcal{D}_{KL}$ \\
        \midrule
        
        GRPO v3.2 & $\tfrac{M_{i,t}}{G |o_i|}$ & $I$ & $I \cdot \mathcal{D}_{KL}$ & 
        Dr GRPO & $\tfrac{1}{G}$ & $I$ & $\mathcal{D}_{KL}$ \\
        \midrule
        
        GSPO & $\alpha_i^S$ & 
        $sg \Big[ \tfrac{\frac{\pi_\theta(y_i|x_i)}{\pi_{\theta_{old}}(y_i|x_i)}^{\frac{1}{|o_i|}}}{\pi_\theta} \Big]\pi_\theta$ & 
        $\times$ &
        GPG & $\tfrac{\hat\alpha}{F_{norm}\sum |o_i|}$ & $\log(\pi_\theta)$ & $\times$\\
        \midrule
        
        GTPO & $\tfrac{\delta_i \lambda_{i,t}}{G |o_i|}$ & $I$ & 
        $\tfrac{1}{G} \sum\nolimits_{i} \tfrac{\delta_i \langle H \rangle_i}{|o_i|} \sum\nolimits_{t} I\lambda_{i,t}$ &
        CISPO & $\tfrac{M_{i,t}}{\sigma(r)\sum |o_i|}$& $sg[I]\log \pi_\theta$ & $\times$\\
        \midrule
        
        DAPO & $\tfrac{1}{\sigma(r)\sum |o_i|}$ & $I$ & $\times$ &
        GCPO & $\tfrac{1}{\sigma(r) G}$ & $\tfrac{\pi_\theta(y_i|x_i)}{\pi_{\theta_{old}}(y_i|x_i)}$ & $\times$ \\
        
        \bottomrule
    \end{tabular}
    }
\end{table*}

\normalsize

\section{Biases in Token-level Gradients}

In this section, we analyze how \grpol objectives affect tokens that are shared across multiple answers. We focus on the initial portion of the generated sequences, where answers are most likely to share identical prefixes and where, due to left-to-right autoregressive generation, updates applied to early tokens have a global effect on the entire sequence. Consider the first $k$ tokens that are identical across a subset of answers. For these positions, the policy probability $\pi_\theta(y_t \mid x, y_{<t})$ is the same for all answers in the group. As a result, the gradient contributions derived from Equation~\ref{eq:grpol}
for these shared tokens differ only through their weighting terms and associated advantages. We formalize the exact form of this aggregate gradient contribution for shared prefixes in the following proposition:

\begin{figure}[h]
\begin{myprop}
\begin{proposition}
\label{prop:first-tokens}
Consider a policy $\pi_\theta$ optimized with Eq. \ref{eq:grpol} via centered advantages (Eq. \ref{eq:advantage}). For any subset of answers $ \tilde{G} \subseteq G$ sharing a common prefix $y_{i, 1:|k|}$, the gradient with respect to this prefix is modulated by the aggregate term $\mathcal{W}_{\text{agg}} = \sum_{i \in \tilde{G}} \omega_i A_i$, where $\omega_i = \alpha_i*s_i(\theta)$.
\end{proposition}
\end{myprop}
\end{figure}

This observation reveals a source of structural bias in token-level gradients. This phenomenon is particularly pronounced when tokens are shared across all sequences. While Eq.~\ref{eq:advantage} implies that the gradient contributions of such tokens
would cancel out under uniform weighting, the actual gradient they receive depends
on the aggregated term $\mathcal{W}_{\text{agg}}$. Consequently, the choice of weighting scheme directly determines how much influence each completion exerts on the shared prefix, independently of the semantic content of the later tokens.
For example, when $\omega_i$ is inversely proportional to the output length,
$\omega_i \propto \frac{1}{|o_i|}$, answers with shorter lengths and positive
advantages contribute disproportionately to the gradient of the initial tokens.
As a consequence, the model is implicitly encouraged to favor shorter outputs,
even when length is not aligned with task quality~\cite{liu2025understanding}.
From an optimization perspective, the induced bias on shared prefix tokens constitutes
a distinct training signal.
Depending on the application, this signal may be exploited, for instance, to control
verbosity, or it may need to be mitigated to avoid unintended stylistic or structural
preferences~\cite{simoni2025gtpo}.

\section{Effects of AdamW Optimizer}

We now turn our attention to the AdamW optimizer~\cite{loshchilov2017decoupled}, the standard choice for GRPO training setups~\cite{simoni2025gtpo, shao2024deepseekmath, yu2025dapo, liu2025understanding}. Analyzing AdamW is particularly relevant in this setting, as the interplay between multiple gradient steps per group and policy clipping significantly alters optimization dynamics. The AdamW update rule is formally defined as follows:
\begin{align}
    \theta_t &= \theta_{t-1} + \xi \frac{\hat{m}_t}{\sqrt{\hat{v}_t} + \epsilon}
    + \xi \lambda \theta_{t-1}
    \label{eq:adam}\\
    m_t &= \frac{\beta_1 m_{t-1}}{{1-\beta_1^{t-1}}}
    + \frac{(1-\beta_1) g_t}{1-\beta_1^{t}}
    \label{eq:m-adam}\\
    v_t &= \frac{\beta_2 v_{t-1}}{{1-\beta_2^{t-1}}}
    + \frac{(1-\beta_2) (g_t)^2}{1-\beta_2^t}
    \label{eq:v-adam}
\end{align}
where $g_t = \nabla_{\theta}\mathcal{J}_{\text{GRPO-L}}(\theta)$ denotes the gradient of
the \grpol objective (the full derivation is reported in Appendix
Eq.~\ref{eq:grad-grpol}).
Unlike standard gradient descent, the update depends not only on the current gradient,
but also on exponentially smoothed estimates of its first- and second-order moments. 

\paragraph{Reward Scaling.}
Despite the extensive literature emphasizing the criticality of reward scaling for stabilizing reinforcement learning algorithms~\cite{reward-scaling-rl, reward-scaling-ppo}, the adaptive nature of AdamW warrants a re-examination of this premise in the context of \grpol algorithm.
We investigate the effect of scaling the reward signal by a factor $\phi \in \mathbb{R}^+$, such that $r_i^* = \phi r_i$. Whether applied to control signal magnitude or induced by normalization, this scaling theoretically alters the optimization landscape. We establish the following property regarding AdamW's response to such transformations when regularization is omitted ($\beta = 0$), a configuration empirically shown to enhance performance in domains like mathematics~\cite{liu2025deepseek}.

\begin{myprop} \begin{proposition} Assume $\beta = 0$ in Eq.~\ref{eq:grpol} and define a scaled reward $r_i^* = \phi r_i$, with $\phi \in \mathbb{R}^+$. In the limit where the numerical stability term $\frac{\epsilon}{\phi \sqrt{\hat{v}_t}} \to 0$, the Adam update in Eq.~\ref{eq:adam} is invariant to the scaling factor $\phi$. \label{prop:scaling-reward} \end{proposition} \end{myprop}

This result, formally derived in the Appendix \ref{app:adam-reward}, shows that without regularization, uniformly scaling the reward does not alter the optimization trajectory
under AdamW.
Intuitively, the adaptive normalization induced by $\hat{v}_t$ compensates for changes
in gradient magnitude, effectively canceling out the effect of reward scaling and
preserving the update direction. However, this invariance no longer holds once a regularization term is introduced
(i.e., $\beta \neq 0$).
In this case, scaling the reward modifies the relative strength between the
reward-driven gradient and the regularization penalty, making the optimization
dynamics explicitly dependent on the reward scale.
As a consequence, the choice of reward normalization becomes a meaningful design
decision in \grpol training. Even when $\beta = 0$, the invariance described in
Proposition~\ref{prop:scaling-reward} relies on the numerical stability constant
$\epsilon$ being negligible compared to $\phi \sqrt{\hat{v}_t}$.
Although $\epsilon$ is typically set to a small value ($10^{-8}$ in PyTorch implementation\footnote{\url{https://docs.pytorch.org/docs/stable/generated/torch.optim.AdamW.html}}), some
reinforcement learning implementations adopt larger values such as $10^{-5}$~\cite{37implementation}.
In these cases, $\epsilon$ may become comparable to small gradient magnitudes,
reintroducing sensitivity to reward scaling.
Despite its potential impact on convergence, the value of $\epsilon$ is often omitted
from reported hyperparameters.

\noindent\paragraph{Adam Overshoot.}

We next analyze the interplay between AdamW and the clipping mechanism in \grpol objectives. This interaction is critical when performing multiple optimization steps on the same batch, where clipping is intended to enforce a trust region. We consider a scenario where the parameter vector reaches the clipping boundary at iteration $T$. We demonstrate that even if the advantage-based gradients vanish at this boundary, the optimizer’s internal dynamics do not cease, driving updates beyond the intended constraints.

\begin{myprop} \begin{proposition} Let $\theta_T$ denote a parameter state at iteration $T$ that lies on the boundary of the clipped region. Even if the instantaneous gradient of the advantage term becomes zero for all $t > T$, the Adam update $\Delta \theta_{T+k}$ continues to move the parameters further into the clipped region. \label{prop:adam-overshoot} \end{proposition} \end{myprop}

The underlying reason is Adam’s momentum mechanism.
Once the parameters enter the clipped region, the gradient contribution of the
advantage term is suppressed by the clipping operation.
However, the first moment estimate retains information from previous gradients and
continues to produce non-zero updates.
As a result, the optimizer keeps moving in the same direction even in the absence of
a corrective gradient signal. For \grpol algorithms, this behavior induces a form of \emph{unidirectional drift}.
If the policy enters an untrusted region during these updates, self-correction becomes impossible. As a result, the model progressively deviates from the trust region until new data is generated in the subsequent iteration. 
\grpol algorithms converge even when clipping is inactive ($\mu = 1$)~\cite{shao2024deepseekmath,simoni2025gtpo,GPG}.This implies the mechanism may be unnecessary, and its complete omission is a promising direction for future work.
The derivation of Proposition~\ref{prop:adam-overshoot} is in Appendix \ref{app:adam-overshoot}.

\section{Conclusion}
In this work, we established a unified formulation for Group Relative Policy Optimization and its variants, revealing disconnects between heuristics and theory. Our analysis identified distinct properties: first, that specific weighting schemes introduce structural gradient biases into shared prefixes; second, the interaction between AdamW momemntum and \grpol objective, in absence of regularization term, makes the objective insensitive to the global reward scaling; and third, that the interaction between AdamW momentum and the objective clipping mechanisms causes parameters to overshoot trust regions, undermining the stability of multi-step updates. 
These findings suggest that the empirical scalability of \grpol methods is achieved at the expense of optimization transparency, necessitating a re-evaluation of current post-training strategies to ensure rigorous alignment between surrogate objectives and desired policy outcomes.


\section*{Limitations}
Our theoretical analysis relies on the assumption of standard autoregressive generation and may not fully generalize to non-standard attention mechanisms or bidirectional architectures. Additionally, while we identified the momentum-induced drift in AdamW, we did not propose a closed-form correction for the optimizer itself, leaving the development of momentum-aware clipping strategies for future work. Finally, our empirical validation of the "overshoot" phenomenon (Proposition 3) focuses on the standard \grpol implementation and may vary under aggressive regularization regimes or alternative optimizer choices such as RMSProp or SGD.

\bibliography{custom}

\begin{thebibliography}{23}
\providecommand{\natexlab}[1]{#1}

\bibitem[{Achiam(2018)}]{achiam2018spinning}
Joshua Achiam. 2018.
\newblock {Spinning Up in Deep Reinforcement Learning}.

\bibitem[{Chen et~al.(2025)Chen, Li, Gong, Jiang, Fei, Yang, Shan, Yu, Wang, Zhu, Xiao, Du, Zhang, Qiao, Zhang, Du, Guo, Chen, Ding, Sun, Li, Jiao, Zhou, Zhang, Ding, Sun, Feng, Cai, Zhu, Sun, Zhuang, Cai, Song, Zhu, Li, Tian, Liu, Xu, Yan, Liu, He, Feng, Yang, Xiao, Han, Wang, Yu, Feng, Li, Zheng, Du, Yang, Zeng, Yu, Tao, Chi, Zhang, Lin, Hu, Di, Gao, Li, Zhao, Ren, Xu, Li, Wang, Tian, Leng, Chen, Chen, Shi, Weng, Guan, Yu, Li, Zhu, Li, Cai, Liang, Cheng, Kong, Li, Chen, Song, Luo, Su, Li, Han, Hou, Lu, Zou, Shen, Gong, Ma, Wang, Shi, Zhong, and Duan}]{CISPO}
Aili Chen, Aonian Li, Bangwei Gong, Binyang Jiang, Bo~Fei, Bo~Yang, Boji Shan, Changqing Yu, Chao Wang, Cheng Zhu, Chengjun Xiao, Chengyu Du, Chi Zhang, Chu Qiao, Chunhao Zhang, Chunhui Du, Congchao Guo, Da~Chen, Deming Ding, and 80 others. 2025.
\newblock \href {https://doi.org/10.48550/ARXIV.2506.13585} {Minimax-m1: Scaling test-time compute efficiently with lightning attention}.
\newblock \emph{CoRR}, abs/2506.13585.

\bibitem[{Chu et~al.(2025)Chu, Huang, Zhang, Wei, and Wang}]{GPG}
Xiangxiang Chu, Hailang Huang, Xiao Zhang, Fei Wei, and Yong Wang. 2025.
\newblock \href {https://doi.org/10.48550/ARXIV.2504.02546} {{GPG:} {A} simple and strong reinforcement learning baseline for model reasoning}.
\newblock \emph{CoRR}, abs/2504.02546.

\bibitem[{Engstrom et~al.(2020)Engstrom, Ilyas, Santurkar, Tsipras, Janoos, Rudolph, and Madry}]{reward-scaling-ppo}
Logan Engstrom, Andrew Ilyas, Shibani Santurkar, Dimitris Tsipras, Firdaus Janoos, Larry Rudolph, and Aleksander Madry. 2020.
\newblock Implementation matters in deep policy gradients: {A} case study on {PPO} and {TRPO}.
\newblock \emph{CoRR}, abs/2005.12729.

\bibitem[{Huang et~al.(2022)Huang, Dossa, Raffin, Kanervisto, and Wang}]{37implementation}
Shengyi Huang, Rousslan Fernand~Julien Dossa, Antonin Raffin, Anssi Kanervisto, and Weixun Wang. 2022.
\newblock The 37 implementation details of proximal policy optimization.
\newblock In \emph{ICLR Blog Track}.

\bibitem[{Ichihara et~al.(2025)Ichihara, Jinnai, Morimura, Sakamoto, Mitsuhashi, and Uchibe}]{ichihara2025mo}
Yuki Ichihara, Yuu Jinnai, Tetsuro Morimura, Mitsuki Sakamoto, Ryota Mitsuhashi, and Eiji Uchibe. 2025.
\newblock Mo-grpo: Mitigating reward hacking of group relative policy optimization on multi-objective problems.
\newblock \emph{arXiv preprint arXiv:2509.22047}.

\bibitem[{Lin et~al.(2025)Lin, Lin, Xie, and Ji}]{lin2025cppo}
Zhihang Lin, Mingbao Lin, Yuan Xie, and Rongrong Ji. 2025.
\newblock Cppo: Accelerating the training of group relative policy optimization-based reasoning models.
\newblock \emph{arXiv preprint arXiv:2503.22342}.

\bibitem[{Liu et~al.(2025{\natexlab{a}})Liu, Mei, Lin, Xue, Wang, Xu, Wu, Zhang, Lin, Dong et~al.}]{liu2025deepseek}
Aixin Liu, Aoxue Mei, Bangcai Lin, Bing Xue, Bingxuan Wang, Bingzheng Xu, Bochao Wu, Bowei Zhang, Chaofan Lin, Chen Dong, and 1 others. 2025{\natexlab{a}}.
\newblock Deepseek-v3. 2: Pushing the frontier of open large language models.
\newblock \emph{arXiv preprint arXiv:2512.02556}.

\bibitem[{Liu et~al.(2025{\natexlab{b}})Liu, Chen, Li, Qi, Pang, Du, Lee, and Lin}]{liu2025understanding}
Zichen Liu, Changyu Chen, Wenjun Li, Penghui Qi, Tianyu Pang, Chao Du, Wee~Sun Lee, and Min Lin. 2025{\natexlab{b}}.
\newblock Understanding r1-zero-like training: A critical perspective.
\newblock \emph{arXiv preprint arXiv:2503.20783}.

\bibitem[{Loshchilov and Hutter(2017)}]{loshchilov2017decoupled}
Ilya Loshchilov and Frank Hutter. 2017.
\newblock Decoupled weight decay regularization.
\newblock \emph{arXiv preprint arXiv:1711.05101}.

\bibitem[{Mroueh et~al.(2025)Mroueh, Dupuis, Belgodere, Nitsure, Rigotti, Greenewald, Navratil, Ross, and Rios}]{mroueh2025revisiting}
Youssef Mroueh, Nicolas Dupuis, Brian Belgodere, Apoorva Nitsure, Mattia Rigotti, Kristjan Greenewald, Jiri Navratil, Jerret Ross, and Jesus Rios. 2025.
\newblock Revisiting group relative policy optimization: Insights into on-policy and off-policy training.
\newblock \emph{arXiv preprint arXiv:2505.22257}.

\bibitem[{Ouyang et~al.(2022)Ouyang, Wu, Jiang, Almeida, Wainwright, Mishkin, Zhang, Agarwal, Slama, Ray, Schulman, Hilton, Kelton, Miller, Simens, Askell, Welinder, Christiano, Leike, and Lowe}]{rlhf}
Long Ouyang, Jeffrey Wu, Xu~Jiang, Diogo Almeida, Carroll~L. Wainwright, Pamela Mishkin, Chong Zhang, Sandhini Agarwal, Katarina Slama, Alex Ray, John Schulman, Jacob Hilton, Fraser Kelton, Luke Miller, Maddie Simens, Amanda Askell, Peter Welinder, Paul~F. Christiano, Jan Leike, and Ryan Lowe. 2022.
\newblock Training language models to follow instructions with human feedback.
\newblock In \emph{Advances in Neural Information Processing Systems 35: Annual Conference on Neural Information Processing Systems 2022, NeurIPS 2022, New Orleans, LA, USA, November 28 - December 9, 2022}.

\bibitem[{Park et~al.(2025)Park, Kim, Kim, Jo, Choi, Cho, and Ryu}]{park2025clip}
Jaesung~R Park, Junsu Kim, Gyeongman Kim, Jinyoung Jo, Sean Choi, Jaewoong Cho, and Ernest~K Ryu. 2025.
\newblock Clip-low increases entropy and clip-high decreases entropy in reinforcement learning of large language models.
\newblock \emph{arXiv preprint arXiv:2509.26114}.

\bibitem[{Schulman et~al.(2017)Schulman, Wolski, Dhariwal, Radford, and Klimov}]{schulman2017proximal}
John Schulman, Filip Wolski, Prafulla Dhariwal, Alec Radford, and Oleg Klimov. 2017.
\newblock Proximal policy optimization algorithms.
\newblock \emph{arXiv preprint arXiv:1707.06347}.

\bibitem[{Shao et~al.(2024)Shao, Wang, Zhu, Xu, Song, Bi, Zhang, Zhang, Li, Wu et~al.}]{shao2024deepseekmath}
Zhihong Shao, Peiyi Wang, Qihao Zhu, Runxin Xu, Junxiao Song, Xiao Bi, Haowei Zhang, Mingchuan Zhang, YK~Li, Yang Wu, and 1 others. 2024.
\newblock Deepseekmath: Pushing the limits of mathematical reasoning in open language models.
\newblock \emph{arXiv preprint arXiv:2402.03300}.

\bibitem[{Simoni et~al.(2025)Simoni, Fontana, Rossolini, and Saracino}]{simoni2025gtpo}
Marco Simoni, Aleksandar Fontana, Giulio Rossolini, and Andrea Saracino. 2025.
\newblock Gtpo: Trajectory-based policy optimization in large language models.
\newblock \emph{arXiv preprint arXiv:2508.03772}.

\bibitem[{van Hasselt et~al.(2016)van Hasselt, Guez, Hessel, Mnih, and Silver}]{reward-scaling-rl}
Hado van Hasselt, Arthur Guez, Matteo Hessel, Volodymyr Mnih, and David Silver. 2016.
\newblock Learning values across many orders of magnitude.
\newblock In \emph{Advances in Neural Information Processing Systems 29: Annual Conference on Neural Information Processing Systems 2016, December 5-10, 2016, Barcelona, Spain}, pages 4287--4295.

\bibitem[{Wang et~al.(2025)Wang, Wang, Fu, Qu, Cheng, Tang, Zhang, and Huo}]{wang2025slow}
Ziyan Wang, Zheng Wang, Jie Fu, Xingwei Qu, Qi~Cheng, Shengpu Tang, Minjia Zhang, and Xiaoming Huo. 2025.
\newblock Slow-fast policy optimization: Reposition-before-update for llm reasoning.
\newblock \emph{arXiv preprint arXiv:2510.04072}.

\bibitem[{Wu and Liu(2025)}]{GCPO}
Hao Wu and Wei Liu. 2025.
\newblock \href {https://doi.org/10.48550/ARXIV.2510.07790} {{GCPO:} when contrast fails, go gold}.
\newblock \emph{CoRR}, abs/2510.07790.

\bibitem[{Xin et~al.(2025)Xin, Liu, Wang, Zhang, Sui, Hu, and Wang}]{xin2025surrogate}
Rihui Xin, Han Liu, Zecheng Wang, Yupeng Zhang, Dianbo Sui, Xiaolin Hu, and Bingning Wang. 2025.
\newblock Surrogate signals from format and length: Reinforcement learning for solving mathematical problems without ground truth answers.
\newblock \emph{arXiv preprint arXiv:2505.19439}.

\bibitem[{Yao et~al.(2023)Yao, Yu, Zhao, Shafran, Griffiths, Cao, and Narasimhan}]{Tree-of-thoughts}
Shunyu Yao, Dian Yu, Jeffrey Zhao, Izhak Shafran, Tom Griffiths, Yuan Cao, and Karthik Narasimhan. 2023.
\newblock Tree of thoughts: Deliberate problem solving with large language models.
\newblock In \emph{Advances in Neural Information Processing Systems 36: Annual Conference on Neural Information Processing Systems 2023, NeurIPS 2023, New Orleans, LA, USA, December 10 - 16, 2023}.

\bibitem[{Yu et~al.(2025)Yu, Zhang, Zhu, Yuan, Zuo, Yue, Dai, Fan, Liu, Liu et~al.}]{yu2025dapo}
Qiying Yu, Zheng Zhang, Ruofei Zhu, Yufeng Yuan, Xiaochen Zuo, Yu~Yue, Weinan Dai, Tiantian Fan, Gaohong Liu, Lingjun Liu, and 1 others. 2025.
\newblock Dapo: An open-source llm reinforcement learning system at scale.
\newblock \emph{arXiv preprint arXiv:2503.14476}.

\bibitem[{Zheng et~al.(2025)Zheng, Liu, Li, Chen, Yu, Gao, Dang, Liu, Men, Yang et~al.}]{zheng2025group}
Chujie Zheng, Shixuan Liu, Mingze Li, Xiong-Hui Chen, Bowen Yu, Chang Gao, Kai Dang, Yuqiong Liu, Rui Men, An~Yang, and 1 others. 2025.
\newblock Group sequence policy optimization.
\newblock \emph{arXiv preprint arXiv:2507.18071}.

\end{thebibliography}

\newpage
\appendix

\section{Inadequacy of the Surrogate Loss as a Performance Proxy}
\label{app:loss_proxy}

This section provides a detailed analysis of why the \grpol surrogate objective suffers from limitations in representing a reliable performance proxy (an intermediate signal intended to estimate the underlying objective).
While the objecitve is well-defined as an optimization signal,
its numerical value does not admit a consistent or monotonic relationship
with reward improvement, even under idealized conditions.
We formalize this limitation in Proposition \ref{prop:loss_proxy} and explicitly characterize the mechanisms
that decouple the surrogate loss from true policy quality.

\begin{myprop}
\begin{proposition}
\label{prop:loss_proxy}
Consider the surrogate objective $\mathcal{J}_{\text{GRPO-L}}(\theta)$
defined in Eq.~\ref{eq:grpol}.
Assume that importance weights are computed with respect to a fixed
reference policy $\pi_{\text{old}}$ sampled at the initial iteration, i.e.,
\[
s_{i,t} \propto
\frac{\pi_\theta(o_{i,t} \mid q, o_{i,<t})}
     {\pi_{\text{old}}(o_{i,t} \mid q, o_{i,<t})}.
\]
Under group-standardized advantages
$\sum_{i=1}^G \mathcal{A}_i = 0$,
the value of $\mathcal{J}_{\text{GRPO-L}}(\theta)$
is an inconsistent proxy for policy performance.
\end{proposition}
\end{myprop}


\paragraph{General form of the objective.}
Ignoring the clipping operation for analytical clarity,
the \grpol surrogate objective can be written as:

\small
\begin{align}
    \mathcal{J}_{\text{GRPO-L}}(\theta) = \mathbb{E}_{q, \{o_i\}} \Bigg[ &\frac{1}{G} \sum_{i=1}^G \left( \mathcal{A}_i \sum_{t=1}^{|o_i|} \omega_{i,t} \rho_{i,t}(\theta) \right) \notag \\
    &- \beta R(\pi_\theta) \Bigg]
\label{eq:proxy_full}
\end{align}

\normalsize

where: $\mathcal{A}_i = r_i - \frac{1}{G}\sum_{j=1}^G r_j$ is the group-centered advantage; $|o_i|$ is the length of the $i$-th completion; $\rho_{i,t}(\theta) = \frac{\pi_\theta(o_{i,t}\mid q, o_{i,<t})}{\pi_{\text{old}}(o_{i,t}\mid q, o_{i,<t})}$ is the token-level importance sampling ratio; $\omega_{i,t}$ aggregates algorithm-specific weighting choices (e.g., $\alpha_{i,t}$, length normalization, masking strategies); $\beta R(\pi_\theta)$ denotes the regularization term.

The central question addressed in this section is whether the scalar value
of $\mathcal{J}_{\text{GRPO-L}}(\theta)$
can be interpreted as a meaningful indicator of training progress
or policy quality.
To answer this question, we analyze two scenarios:
(A) the first optimization step, where the current policy coincides
with the sampling policy, and
(B) later iterations, where the two policies diverge.

\subsection{Scenario A: First optimization step ($\rho_{i,t}(\theta) = 1$)}

At the first update, the policy has not yet changed,
so $\pi_\theta = \pi_{\text{old}}$ and therefore
$\rho_{i,t}(\theta) = 1$ for all $i,t$.
In this case, all importance sampling effects vanish.

We can absorb the remaining per-token design choices
into a single effective weight $\tilde{\omega}_{i,t}$.
The objective simplifies to:
\begin{equation}
\mathcal{J}_{\text{align}}(\theta)
=
\mathbb{E}
\Bigg[
\sum_{i=1}^G
\mathcal{A}_i\,\Omega_i
-
\beta R(\pi_\theta)
\Bigg] \label{eq:proxy_align}
\end{equation}
where
\[
\Omega_i = \sum_{t=1}^{|o_i|} \tilde{\omega}_{i,t}
\]
is the cumulative weight assigned to trajectory $i$.

This formulation makes explicit that the surrogate objective
depends only on the interaction between advantages $\mathcal{A}_i$
and cumulative weights $\Omega_i$.
We now examine three representative weighting regimes.

\subsubsection*{Case 1: Length-normalized weights}

Many \grpol methods normalize updates by sequence length,
using weights of the form $\tilde{\omega}_{i,t} = \frac{C}{|o_i|}$.
In this case,
\[
\Omega_i = \sum_{t=1}^{|o_i|} \frac{C}{|o_i|} = C,
\]
which is constant across all trajectories. Substituting into Eq.~\ref{eq:proxy_align} yields:
\begin{align}
\mathcal{J}_{\text{align}}(\theta)
&=
\mathbb{E}
\Bigg[
C
\underbrace{\sum_{i=1}^G \mathcal{A}_i}_{=\,0}
-
\beta R(\pi_\theta)
\Bigg]
\notag \\
&=
-\beta\,\mathbb{E}[R(\pi_\theta)].
\end{align}

Thus, the entire reward-driven component of the objective cancels out.
The surrogate loss is fully dominated by the regularization term and
contains no information about relative reward improvement.
In this regime, the loss value is fundamentally uninformative
as a measure of policy performance.

\subsubsection*{Case 2: Constant token-wise weights}

If weights are constant per token,
$\tilde{\omega}_{i,t} = C$ (e.g., Dr.~GRPO),
then the cumulative weight scales linearly with output length:
\[
\Omega_i = C\,|o_i|.
\]

The objective becomes:
\begin{equation}
\mathcal{J}_{\text{align}}(\theta)
=
\mathbb{E}
\Bigg[
C
\sum_{i=1}^G
\mathcal{A}_i |o_i|
-
\beta R(\pi_\theta)
\Bigg]
\end{equation}

In this case, the loss no longer cancels,
but its sign and magnitude reflect whether
positively advantaged completions tend to be longer or shorter
than negatively advantaged ones.
The objective therefore acts as a proxy for sequence length statistics,
not for reward maximization or task correctness.

\subsubsection*{Case 3: General parametric weighting}

More complex methods (e.g., GTPO, CPPO) define
$\tilde{\omega}_{i,t}$ as a non-trivial function of $i$ and $t$.
Here, the reward-weighted sum does not vanish,
but instead satisfies:
\begin{equation}
\mathcal{J}_{\text{align}}(\theta)
\propto
\operatorname{Cov}(\mathcal{A}, \Omega)    
\end{equation}

Although the loss is non-zero,
its value is entirely determined by the interaction between
the advantage distribution and the chosen weighting scheme.
Unless the weights are explicitly designed to encode
task-relevant structure,
the loss magnitude is an artifact of hyperparameterization,
not a measure of learning progress.

\paragraph{Conclusion of Scenario A.}
Across all weighting regimes,
the surrogate loss fails to maintain a consistent or monotonic relationship
with true policy quality.
Its numerical value is therefore an unreliable indicator of performance,
even in the absence of importance sampling effects.

\subsection{Scenario B: Multiple optimization steps ($\rho_{i,t}(\theta) \neq 1$)}

After the first update,
$\pi_\theta$ diverges from $\pi_{\text{old}}$ and
importance sampling ratios $\rho_{i,t}(\theta) \neq 1$ appear.
While this breaks the exact cancellations observed in Scenario~A,
it does not restore interpretability.

The loss value now depends on two independent sources of variability: the structural biases induced by the weighting scheme $\tilde{\omega}_{i,t}$ and stochastic fluctuations of the importance ratios $\rho_{i,t}(\theta)$.

As a result, changes in the surrogate loss primarily reflect
off-policy drift and optimizer dynamics,
rather than genuine reward improvement.
A decreasing loss does not imply better policies,
nor does a stable loss indicate convergence.


\section{Derivation of the Gradient for $\mathcal{J}_{\text{GRPO-L}}(\theta)$}
\label{app:gradient}

This appendix derives the gradient of the \grpol surrogate objective and
makes explicit the token-level structure that later induces shared-prefix biases.
For clarity, we derive the gradient in the region where the \emph{unclipped} term
is active; when the clipped branch is active, the gradient through the advantage
term is zero (up to boundary measure-zero cases).

\subsection{Gradient of the \grpol objective}

Recall the \grpol objective in Eq.~\ref{eq:grpol}.

\small
\begin{align}
\mathcal{J}_{\text{GRPO-L}}(\theta)
&=
\mathbb{E}_{q,\{o_i\}}
\Bigg[
\sum_{i=1}^G
\sum_{t=1}^{|o_i|}
\alpha_{i,t}\,
\min\!\Big(
s_{i,t}(\theta)\,A_i,\;\notag\\
&
\text{clip}(s_{i,t}(\theta), 1-\epsilon_{\text{low}}, 1+\epsilon_{\text{up}})\,A_i
\Big)
-\beta R(\theta)
\Bigg]\notag
\end{align}

\normalsize
We define the token-level importance ratio as
\begin{align}
s_{i,t}(\theta)
&:=k_{i,t} \cdot
\pi_\theta\!\left(y_{i,t}\mid x, y_{i,<t}\right)
\label{eq:ratio_def_app}
\end{align}
Here, $k_{i,t}$ is a term that depends on i and t, but is independent of $\theta$. We now apply the gradient and move $\nabla_\theta$ inside expectation and sums. By linearity,
\begin{align}
\nabla_{\theta}\mathcal{J}_{\text{GRPO-L}}(\theta)
&=
\mathbb{E}_{q,\{o_i\}}
\Bigg[
\sum_{i=1}^G
\sum_{t=1}^{|o_i|}
\alpha_{i,t}\,
\nabla_\theta\,\min(\cdot) \notag\\ &
\;-\;
\beta \nabla_\theta R(\theta)
\Bigg]
\label{eq:grad_step1_split}
\end{align}

The gradient depends on the active branch.
When the unclipped term is active,
\begin{align}
\nabla_\theta \min(\cdot)
&=
\nabla_\theta\!\left(s_{i,t}(\theta)A_i\right)
=
A_i\,\nabla_\theta s_{i,t}(\theta) \notag
\end{align}
while when the clipped term is active, its value is constant w.r.t.\ $\theta$
in the interior of the clipped region, hence the advantage gradient is zero
(ignoring boundary non-differentiability).

Since $\pi_{\theta_{\text{old}}}$ does not depend on current $\theta$,

\small
\begin{align}
\nabla_\theta s_{i,t}(\theta)
&=\nabla_\theta \left(k_{i,t} \cdot
\pi_\theta\!\left(y_{i,t}\mid x, y_{i,<t}\right)\right)
\notag\\
&= k_{i,t} \cdot
\nabla_\theta
\pi_\theta\!\left(y_{i,t}\mid x, y_{i,<t}\right)
\notag
\end{align}

\normalsize
Using the log-derivative trick,
$\nabla_\theta \pi_\theta = \pi_\theta \nabla_\theta \log \pi_\theta$,
we get
\begin{align}
\nabla_\theta s_{i,t}(\theta)
&=\left[k_{i,t} \cdot\pi_\theta\!\left(y_{i,t}\mid x, y_{i,<t}\right) \right]
\cdot \notag\\ &
\cdot\nabla_\theta \log \pi_\theta\!\left(y_{i,t}\mid x, y_{i,<t}\right)
\notag\\
&=
s_{i,t}(\theta)\;
\nabla_\theta \log \pi_\theta\!\left(y_{i,t}\mid x, y_{i,<t}\right).
\label{eq:ratio_grad_final_split}
\end{align}

\normalsize
Substituting Eq.~\ref{eq:ratio_grad_final_split} into
Eq.~\ref{eq:grad_step1_split} yields:
\begin{align}
\nabla_{\theta}\mathcal{J}_{\text{GRPO-L}}&(\theta)
=
\mathbb{E}_{q,\{o_i\}}
\Bigg[
\sum_{i=1}^G
A_i
\sum_{t=1}^{|o_i|}
\alpha_{i,t}\,s_{i,t}(\theta)\,\notag\\
&
\nabla_{\theta}\log\pi_\theta\!\left(y_{i,t}\mid x, y_{i,<t}\right)
-\beta\nabla_{\theta}R(\theta)
\Bigg]
\label{eq:grad-grpol}
\end{align}

\subsection{First token issues}
\label{app:first-token}

We now isolate the gradient contribution on tokens that belong to a prefix shared
by multiple completions in the same group.
Let $|k|$ be the length of a prefix shared by a subset of $\tilde{G}\leq G$
completions.

\paragraph{Prefix/deviation decomposition.}
Splitting the inner sum over time gives:

\small
\begin{align}
\nabla_{\theta}&\mathcal{J}_{\text{GRPO-L}}(\theta)
= \notag\\
&
\mathbb{E}_{q,\{o_i\}}
\Bigg[
\sum_{i=1}^G A_i
\Bigg(
\sum_{t=1}^{|k|}
\alpha_{i,t}\,s_{i,t}(\theta)\,
\nabla_{\theta}\log\pi_\theta\!\left(y_{i,t}\mid x, y_{i,<t}\right)
\notag\\
&
+
\sum_{t=|k|+1}^{|o_i|}
\alpha_{i,t}\,s_{i,t}(\theta)\,
\nabla_{\theta}\log\pi_\theta\!\left(y_{i,t}\mid x, y_{i,<t}\right)
\Bigg)
\notag\\
&\qquad\qquad
-\beta\nabla_{\theta}R(\theta)
\Bigg]
\label{eq:grad-first-token-splitted}
\end{align}
\normalsize

For all $t\le |k|$ and all completions $i$ in the subset that shares the prefix,
both $y_{i,t}$ and its context $y_{i,<t}$ are identical. Hence,

\begin{align}
\nabla_{\theta}\log\pi_\theta\!\left(y_{i,t}\mid x, y_{i,<t}\right)
&=
\nabla_{\theta}\log\pi_\theta\!\left(y_{t}\mid x, y_{<t}\right),
\notag\\
&\forall\, i\in\{1,\dots,\tilde{G}\},\; t\le |k|
\label{eq:shared_prefix_scorefn}
\end{align}

Define the aggregated coefficient
\begin{align}
\omega_{i,t}
\;:=\;
\alpha_{i,t}\,s_{i,t}(\theta)
\label{eq:omega_def_split}
\end{align}

Then the gradient restricted to the shared prefix
(denoted $\nabla_\theta \tilde{\mathcal{J}}_{\text{GRPO-L}}(\theta)$) becomes:

\small
\begin{align}
\nabla_{\theta}\tilde{\mathcal{J}}_{\text{GRPO-L}}(\theta)
&=
\sum_{t=1}^{|k|}
\nabla_{\theta}\log\pi_\theta\!\left(y_{t}\mid x, y_{<t}\right)\;
\sum_{i=1}^{\tilde{G}} A_i\,\omega_{i,t}\notag
\end{align}
\normalsize

In the following, when it does not change the qualitative argument,
we suppress the explicit dependence on $t$ and write $\omega_i$ for simplicity.

\subsubsection*{Case 1: Constant token-wise weights}

Assume uniform weights over completions: $\omega_i=C$.
Then:

\small

\begin{align}
\nabla_{\theta}\tilde{\mathcal{J}}_{\text{GRPO-L}}(\theta)
&=
C
\sum_{t=1}^{|k|}
\nabla_{\theta}\log\pi_\theta\!\left(y_{t}\mid x, y_{<t}\right)\;
\sum_{i=1}^{\tilde{G}} A_i\notag
\end{align}

\normalsize

Since $A_i = R_i - \bar{R}$ is group-centered,
the behavior depends on which completions share the prefix:
(i) if the prefix occurs only in $A_i>0$ completions, it is reinforced;
(ii) in mixed regimes, the net update is the algebraic sum;
(iii) if the prefix is ubiquitous across all $G$ completions,
$\sum_{i=1}^{G}A_i=0$ and the update cancels.

\subsubsection*{Case 2: Non-uniform weighting over $i$}

If weights depend on the completion index, $\omega_i\neq \text{const}$, then:

\small

\begin{align}
\nabla_{\theta}\tilde{\mathcal{J}}_{\text{GRPO-L}}(\theta)
&=
\sum_{t=1}^{|k|}
\nabla_{\theta}\log\pi_\theta\!\left(y_{t}\mid x, y_{<t}\right)\;
\sum_{i=1}^{\tilde{G}} \omega_i\,A_i\notag
\end{align}

\normalsize

In this regime, cancellations generally do not hold:
shared-prefix tokens can receive a net update dominated by the completions
with larger $\omega_i$, which can induce systematic biases unrelated to
semantic quality (e.g., length preferences when $\omega_i$ depends on $|o_i|$).

\section{Reward magnitude and Adam}
\label{app:adam-reward}

This section analyzes how scaling the reward signal affects \grpol training
when optimization is performed with Adam/AdamW.
We first show that group-centered advantages scale linearly with the reward.
We then propagate this scaling through (i) the \grpol gradient,
(ii) Adam's first and second moments, and (iii) the final parameter update.
The key takeaway is that, when the regularization term is absent (or negligible),
Adam is approximately invariant to global reward scaling.

\subsection{Scaling properties of the advantage term}

We start by characterizing how the \grpol advantage behaves under a linear
transformation of the reward.

\begin{myprop}
\begin{proposition}[Advantage scaling]
\label{prop:adv_scaling}
Let the group-centered advantage be
$
A_i = R_i -  \frac{1}{G}\sum_{j=1}^G R_j.
$
If rewards are scaled by a constant $\phi \in \mathbb{R}$,
$
R_i^* = \phi R_i,
$
then the transformed advantage satisfies
\begin{equation}
A_i^* = \phi A_i
\end{equation}
\end{proposition}
\end{myprop}

\begin{proof}
First compute the transformed group baseline:

\small

\begin{align}
\bar{R}^*
&= \frac{1}{G}\sum_{j=1}^G R_j^*
= \frac{1}{G}\sum_{j=1}^G \phi R_j
= \phi \Bigg(\frac{1}{G}\sum_{j=1}^G R_j\Bigg)
= \phi \bar{R} \notag
\end{align}

\normalsize

Then the transformed advantage is
\begin{align}
A_i^*
&= R_i^* - \bar{R}^*
= \phi R_i - \phi \bar{R}
= \phi (R_i - \bar{R})
= \phi A_i
\end{align}
\end{proof}

\subsection{Gradient decomposition and scaling properties}

We decompose the \grpol gradient into an advantage-driven term and a
regularization term. Using Eq.~\ref{eq:grad-grpol}, define:

\small
\begin{align}
g_A(\theta)
& :=
\sum_{i=1}^G
A_i
\sum_{t=1}^{|o_i|}
\alpha_{i,t}\,s_{i,t}(\theta)\,
\nabla_{\theta}\log\pi_\theta\!\left(y_{i,t}\mid x, y_{i,<t}\right)
\label{eq:gA_def}
\\[2pt]
g_R(\theta)
& :=
\beta \nabla_\theta R(\theta)
\label{eq:gR_def}
\end{align}
\normalsize

so that the total gradient is $g(\theta)=g_A(\theta)-g_R(\theta)$.

By Proposition~\ref{prop:adv_scaling}, scaling the rewards by $\phi$ implies
$A_i^*=\phi A_i$. Therefore the advantage-driven component scales linearly:

\small
\begin{align}
g_A^*(\theta)
&=
\sum_{i=1}^G
A_i^*
\sum_{t=1}^{|o_i|}
\alpha_{i,t}\,s_{i,t}(\theta)\,
\nabla_{\theta}\log\pi_\theta\!\left(y_{i,t}\mid x, y_{i,<t}\right)
\notag\\
&=
\phi
\sum_{i=1}^G
A_i
\sum_{t=1}^{|o_i|}
\alpha_{i,t}\,s_{i,t}(\theta)\,
\nabla_{\theta}\log\pi_\theta\!\left(y_{i,t}\mid x, y_{i,<t}\right)=\notag\\
&=\phi\, g_A(\theta)
\label{eq:gA_scaling}
\end{align}

\normalsize

Conversely, $g_R(\theta)$ is unaffected by reward scaling because it depends only
on the regularizer and $\beta$.

\subsection{Adam moments under gradient scaling}

We now study how Adam's moments scale when the gradient is multiplied by $\phi$.
Let $g_t$ be the gradient at optimization step $t$, and assume
\begin{align}
g_t^* &= \phi g_t
\qquad \text{for all } t\ge 1
\notag
\end{align}

Adam maintains exponential moving averages:
\begin{align}
m_t &= \beta_1 m_{t-1} + (1-\beta_1) g_t
\notag
\\
v_t &= \beta_2 v_{t-1} + (1-\beta_2) g_t^{ 2}
\notag
\end{align}
with bias-corrected versions
\begin{align}
\hat m_t &= \frac{m_t}{1-\beta_1^t},
&
\hat v_t &= \frac{v_t}{1-\beta_2^t}.
\notag
\end{align}

\begin{myprop}
\begin{proposition}[Moment scaling]
\label{prop:adam_moments_scaling}
If $g_t^*=\phi g_t$ for all $t$, then for all $t\ge 1$:
\begin{align}
m_t^* &= \phi m_t,
&
v_t^* &= \phi^2 v_t,
\end{align}
and equivalently $\hat m_t^*=\phi \hat m_t$ and $\hat v_t^*=\phi^2 \hat v_t$.
\end{proposition}
\end{myprop}

\begin{proof}
We prove by induction.

\textit{Base case ($t=1$).} With $m_0=v_0=0$,
\begin{align}
m_1^*
&= (1-\beta_1) g_1^*
= (1-\beta_1)\phi g_1
= \phi m_1 \notag
\\
v_1^*
&= (1-\beta_2) (g_1^*)^{ 2}
= (1-\beta_2)\phi^2 g_1^{ 2}
= \phi^2 v_1 \notag
\end{align}

\textit{Inductive step.} Assume $m_{t-1}^*=\phi m_{t-1}$ and $v_{t-1}^*=\phi^2 v_{t-1}$.
Then
\begin{align}
m_t^*
&= \beta_1 m_{t-1}^* + (1-\beta_1) g_t^*
\notag\\
&= \beta_1 (\phi m_{t-1}) + (1-\beta_1)(\phi g_t)=\notag\\
&= \phi\big(\beta_1 m_{t-1} + (1-\beta_1) g_t\big)
= \phi m_t 
\\\notag\\
v_t^*
&= \beta_2 v_{t-1}^* + (1-\beta_2) (g_t^*)^{ 2}
\notag\\
&= \beta_2 (\phi^2 v_{t-1}) + (1-\beta_2)\phi^2 g_t^{ 2}=\notag\\
&= \phi^2\big(\beta_2 v_{t-1} + (1-\beta_2) g_t^{ 2}\big)
= \phi^2 v_t 
\end{align}
Bias correction divides by $(1-\beta_1^t)$ and $(1-\beta_2^t)$, hence it preserves
the same scaling. 
\end{proof}

\subsection{Adam update invariance under reward scaling}
\label{app:adam_invariance}

We now analyze when Adam becomes invariant to global reward scaling.
Assume the regularization term is absent or negligible, i.e., $g_R(\theta)\approx 0$.
Then $g_t$ is driven only by the advantage term and scales as $g_t^*=\phi g_t$.

AdamW updates parameters as
\begin{align}
\Delta\theta_t
&=
-\xi \frac{\hat m_t}{\sqrt{\hat v_t}+\epsilon}
\;-\;
\xi\lambda \theta_{t-1}\notag
\end{align}
where $\xi$ is the learning rate, $\epsilon$ the numerical stabilizer,
and $\lambda$ the weight decay coefficient.

Using Proposition~\ref{prop:adam_moments_scaling}, we have
$\hat m_t^*=\phi \hat m_t$ and $\hat v_t^*=\phi^2 \hat v_t$, hence
\begin{align}
\Delta\theta_t^*
&=
-\xi \frac{\hat m_t^*}{\sqrt{\hat v_t^*}+\epsilon}
\;-\;
\xi\lambda \theta_{t-1}
\notag\\
&=
-\xi \frac{\phi \hat m_t}{\sqrt{\phi^2 \hat v_t}+\epsilon}
\;-\;
\xi\lambda \theta_{t-1}
=\notag\\
&=-\xi \frac{\phi \hat m_t}{\phi\sqrt{\hat v_t}+\epsilon}
\;-\;
\xi\lambda \theta_{t-1}\notag
\end{align}

Factor $\phi$ out of the denominator (assuming $\phi>0$):
\begin{align}
\Delta\theta_t^*
&=
-\xi \frac{\hat m_t}{\sqrt{\hat v_t}\Big(1+\frac{\epsilon}{\phi\sqrt{\hat v_t}}\Big)}
\;-\;
\xi\lambda \theta_{t-1}\notag
\end{align}

Therefore, in the regime where $\epsilon \ll \phi\sqrt{\hat v_t}$, we obtain the
approximate invariance:
\begin{align}
\lim_{\frac{\epsilon}{\phi\sqrt{\hat v_t}}\to 0}
\Delta\theta_t^*
&=
-\xi \frac{\hat m_t}{\sqrt{\hat v_t}}
\;-\;
\xi\lambda \theta_{t-1}
=
\Delta\theta_t.
\label{eq:adam_invariance_limit}
\end{align}

This shows that when the optimization signal is purely reward-driven,
Adam's adaptive normalization cancels global reward scaling.
However, if a regularization term is present ($\beta\neq 0$), then
the total gradient becomes $g_t = g_{A,t} - g_{R,t}$ and scaling the rewards
changes the relative strength between the two components, breaking invariance.

\section{Adam overly moves your model}
\label{app:adam-overshoot}

This section analyzes the interaction between \grpol clipping and Adam's
momentum. The key point is that clipping can zero out the instantaneous
advantage gradient once the policy ratio exits the trust region, but Adam's
first-moment accumulator can continue to move parameters in the same direction,
causing overshoot into the clipped region.

\subsection{Gradient discontinuity induced by clipping}

Let $\mathcal{R}_{\text{clip}}$ denote the subset of parameter space where the
importance ratio exceeds the clip bounds in the direction favored by $A_i$
(e.g., $s_{i,t}>1+\epsilon_{\text{up}}$ with $A_i>0$, or
$s_{i,t}<1-\epsilon_{\text{low}}$ with $A_i<0$).
Inside this region, the advantage term is clipped and its gradient is zero.

Equivalently, the gradient takes the piecewise form:

\small
\begin{align}
\nabla_\theta \mathcal{J}_{\text{GRPO-L}}(\theta)
=
\begin{cases}
\nabla_\theta \mathcal{J}_{\text{ADV}}(\theta) - \beta \nabla_\theta R(\theta),
& \theta \notin \mathcal{R}_{\text{clip}},
\\[4pt]
-\beta \nabla_\theta R(\theta),
& \theta \in \mathcal{R}_{\text{clip}}.
\end{cases}
\label{eq:grad_discontinuity}
\end{align}

\normalsize

Intuitively, if $\beta$ is small, entering $\mathcal{R}_{\text{clip}}$ should
dramatically reduce the gradient magnitude and stop motion in that direction.
The next subsection shows why Adam can violate this intuition.

\subsection{Proposition: momentum overshoot}

\begin{myprop}
\begin{proposition}[Momentum overshoot]
\label{prop:adam_overshoot}
Let $\theta_T$ be a parameter iterate lying on the boundary of the clipped region.
Assume that for $t>T$ the advantage gradient becomes zero due to clipping,
i.e., $g_{A,t}=0$. Then, even if the instantaneous advantage gradient remains
zero for subsequent inner-loop steps, Adam can continue to update parameters
in the same direction, pushing the iterate deeper into $\mathcal{R}_{\text{clip}}$.
\end{proposition}
\end{myprop}

\begin{proof}
For steps $t<T$, assume the advantage gradient points consistently toward the
upper clip boundary, i.e., $g_{A,t}$ has a persistent sign that increases
$s_{i,t}(\theta)$.
Adam accumulates these gradients in the first moment:
\begin{align}
m_t
&= \beta_1 m_{t-1} + (1-\beta_1) g_t \notag
\end{align}

At $t=T$, the iterate enters $\mathcal{R}_{\text{clip}}$ and the advantage
gradient is suppressed: $g_{A,T}=0$ (and similarly for all $t>T$).
Neglecting regularization for exposition, the new first-moment update becomes:
\begin{align}
m_T
&= \beta_1 m_{T-1} + (1-\beta_1)\underbrace{g_T}_{=\,0}
= \beta_1 m_{T-1}\notag
\\
m_{T+k}
&= \beta_1^{k+1} m_{T-1},
\qquad k\ge 0.
\notag
\end{align}

Thus, even though the instantaneous gradient is zero, $m_{T+k}$ remains non-zero
for many steps when $\beta_1$ is close to one (e.g., $\beta_1=0.9$).
Since the Adam update depends on $\hat m_t$, the parameter update remains non-zero:
\begin{align}
\Delta\theta_{T+k}
&=
-\xi \frac{\hat m_{T+k}}{\sqrt{\hat v_{T+k}}+\epsilon}
\;-\;
\xi\lambda \theta_{T+k-1}
\label{eq:adam_update_inside_clip}
\end{align}

Therefore, the iterate continues to move in the direction encoded by the
pre-clipping momentum, pushing the ratio further beyond the clip boundary.
Clipping acts as a ``hard stop'' for the instantaneous gradient, but Adam's
momentum makes it a ``soft brake'' for the parameter trajectory.
\end{proof}

\paragraph{Practical implication.}
When multiple optimization steps are applied on the same sampled group (inner loop),
the overshoot effect becomes more pronounced: the policy can drift further into
the clipped region before new samples are generated, weakening the intended trust-region
interpretation of clipping.

\paragraph{Quantifying overshoot (Adam canonical form).}
We now quantify how large the Adam step can remain \emph{after} entering the clipped
region, even when the instantaneous advantage gradient becomes zero.

Assume that at step $T$ the iterate enters $\mathcal{R}_{\text{clip}}$, so that
the advantage gradient is suppressed for all subsequent inner-loop steps,
i.e., $g_{A,t}=0$ for $t\ge T$. For clarity, we first ignore weight decay and
regularization and focus on the Adam preconditioned direction
$\hat m_t / (\sqrt{\hat v_t}+\epsilon)$.

Under $g_T=0$, Adam moment recurrences reduce to pure exponential decay:

\small

\begin{align}
m_T
&=
\beta_1 m_{T-1} + (1-\beta_1)\underbrace{g_T}_{0}
=
\beta_1 m_{T-1} \notag
\\
v_T
&=
\beta_2 v_{T-1} + (1-\beta_2)\underbrace{g_T^{ 2}}_{0}
=
\beta_2 v_{T-1}\notag
\end{align}

\normalsize
Using bias correction,

\small
\begin{align}
\hat m_T
&=
\frac{m_T}{1-\beta_1^T}
=
\frac{\beta_1 m_{T-1}}{1-\beta_1^T},
&
\hat v_T
&=
\frac{v_T}{1-\beta_2^T}
=
\frac{\beta_2 v_{T-1}}{1-\beta_2^T}\label{eq:adam_biascorr_at_entry}
\end{align}

\normalsize

Similarly,
\begin{align}
\hat m_{T-1}
&=
\frac{m_{T-1}}{1-\beta_1^{T-1}},
&
\hat v_{T-1}
&=
\frac{v_{T-1}}{1-\beta_2^{T-1}} \notag
\end{align}

\medskip
\noindent\textbf{A $C_T$-like coefficient.}
Define the ratio between the (magnitude of the) preconditioned update
immediately after clipping and the one immediately before clipping:
\begin{align}
C_T
&:=
\frac{
\left\lVert \dfrac{\hat m_T}{\sqrt{\hat v_T}+\epsilon} \right\rVert
}{
\left\lVert \dfrac{\hat m_{T-1}}{\sqrt{\hat v_{T-1}}+\epsilon} \right\rVert
}.
\label{eq:CT_def}
\end{align}
In the common regime where $\epsilon \ll \sqrt{\hat v_{T-1}}$ and
$\epsilon \ll \sqrt{\hat v_{T}}$, we can approximate
\begin{align}
C_T
&\approx
\frac{
\left\lVert \dfrac{\hat m_T}{\sqrt{\hat v_T}} \right\rVert
}{
\left\lVert \dfrac{\hat m_{T-1}}{\sqrt{\hat v_{T-1}}} \right\rVert
}
=
\frac{\left\lVert \hat m_T \right\rVert}{\left\lVert \hat m_{T-1} \right\rVert}
\cdot
\frac{\sqrt{\hat v_{T-1}}}{\sqrt{\hat v_{T}}}\notag
\end{align}
Substituting Eq.~\ref{eq:adam_biascorr_at_entry} yields
\begin{align}
C_T
&\approx
\Bigg[
\beta_1\,
\frac{1-\beta_1^{T-1}}{1-\beta_1^{T}}
\Bigg]
\cdot
\Bigg[
\sqrt{
\frac{1}{\beta_2}\,
\frac{1-\beta_2^{T}}{1-\beta_2^{T-1}}
}
\Bigg].
\label{eq:CT_closed_form}
\end{align}

This coefficient captures how much ``inertia'' remains \emph{exactly at the first
step} after the advantage gradient is clipped out.
In the limit $T\to\infty$, bias-correction saturates and we obtain:
\begin{align}
\lim_{T\to\infty} C_T
&=
\frac{\beta_1}{\sqrt{\beta_2}}.
\label{eq:CT_asymptotic}
\end{align}
For typical values $(\beta_1,\beta_2)=(0.9,0.95)$,
\begin{align}
\frac{\beta_1}{\sqrt{\beta_2}}
&=
\frac{0.9}{\sqrt{0.95}}
\approx 0.923 \notag
\end{align}
meaning that the \emph{first} post-clipping step can still be on the order of
$\sim 92\%$ of the previous preconditioned step once training is past the early
bias-correction transient.

\medskip
\noindent\textbf{Overshoot across $k$ inner-loop steps.}
The same reasoning extends to subsequent clipped steps. For $k\ge 0$, when
$g_{T+k}=0$ we have
\begin{align}
m_{T+k}
&=
\beta_1^{k+1} m_{T-1},
&
v_{T+k}
&=
\beta_2^{k+1} v_{T-1}.
\label{eq:adam_decay_k}
\end{align}
Define an extension of Eq.~\ref{eq:CT_def}:
\begin{align}
C_{T,k}
&:=
\frac{
\left\lVert \dfrac{\hat m_{T+k}}{\sqrt{\hat v_{T+k}}+\epsilon} \right\rVert
}{
\left\lVert \dfrac{\hat m_{T-1}}{\sqrt{\hat v_{T-1}}+\epsilon} \right\rVert
}.
\end{align}
Again for $\epsilon$ negligible, we obtain the closed form
\begin{align}
C_{T,k}
&\approx
\Bigg[
\beta_1^{k+1}\,
\frac{1-\beta_1^{T-1}}{1-\beta_1^{T+k}}
\Bigg]
\cdot
\Bigg[
\sqrt{
\frac{1}{\beta_2^{k+1}}\,
\frac{1-\beta_2^{T+k}}{1-\beta_2^{T-1}}
}
\Bigg].
\label{eq:CTk_closed_form}
\end{align}
For large $T$ (where bias correction is stable), Eq.~\ref{eq:CTk_closed_form}
simplifies to an exponential decay:
\begin{align}
C_{T,k}
&\approx
\left(\frac{\beta_1}{\sqrt{\beta_2}}\right)^{k+1}.
\label{eq:CTk_asymptotic}
\end{align}
With $(\beta_1,\beta_2)=(0.9,0.95)$ this gives
$C_{T,4}\approx 0.923^{5}\approx 0.66$,
i.e., even after \emph{five} clipped inner-loop steps the update magnitude can
still be around $\sim 66\%$ of the pre-clipping step, which explains why the
policy can drift substantially deeper into the clipped region before new samples
are generated.

\medskip
\noindent\textbf{Effect of $\epsilon$.}
When $\epsilon$ is not negligible (e.g., for very small $\hat v_t$), the ratios
in Eq.~\ref{eq:CT_closed_form}--\ref{eq:CTk_closed_form} are further modulated by
\begin{align}
\frac{\sqrt{\hat v_{T-1}}+\epsilon}{\sqrt{\hat v_{T+k}}+\epsilon},
\end{align}
which can either dampen or amplify the residual step depending on the scale of
$\hat v_t$ relative to $\epsilon$.

\end{document}